\documentclass{article} 
\usepackage{iclr2025_conference,times}

\usepackage{dsfont}

\usepackage{hyperref}
\usepackage{url}
\usepackage{acro}
\usepackage{amsmath}
\usepackage{graphicx}
\usepackage{booktabs}
\usepackage{multirow}
\usepackage{colortbl}
\usepackage{makecell}
\usepackage{subcaption}

\usepackage[capitalize]{cleveref}
\crefname{section}{Sec.}{Secs.}
\Crefname{section}{Section}{Sections}
\Crefname{table}{Table}{Tables}
\crefname{table}{Tab.}{Tabs.}
\Crefname{algorithm}{Algorithm}{Algorithms}
\crefname{algorithm}{Alg.}{Algs.}
\Crefname{appendix}{Appendix}{Appendices}
\crefname{appendix}{App.}{Apps.}

\newcommand{\greyrule}{\arrayrulecolor{black!15}\midrule\arrayrulecolor{black}}

\newcommand{\grtext}{\color{gray}}      %

\usepackage{amssymb}
\usepackage{pifont}
\newcommand{\xmark}{\ding{55}}%

\makeatletter
\renewcommand{\paragraph}[1]{\vspace{0.0pt}{\bf #1}}
\makeatother

\newcommand{\tablesize}{\fontsize{8}{9}\selectfont} 

\newcommand{\myparagraph}[1]{\vspace{0.0pt}\noindent{\bf #1}}

\DeclareAcronym{mot}{
  short = MOT ,
  long = multiple object tracking
}
\DeclareAcronym{reid}{
  short = Re-ID ,
  long = re-identification
}
\DeclareAcronym{nms}{
  short = NMS ,
  long = non maximum suppression
}
\DeclareAcronym{iou}{
  short = IoU ,
  long = Intersection over Union 
}
\DeclareAcronym{roi}{
  short = RoI ,
  long = region of interest ,
  long-plural-form = regions of interest
}
\DeclareAcronym{deta}{
  short = DetA ,
  long = detection accuracy ,
}
\DeclareAcronym{assa}{
  short = AssA ,
  long = association accuracy ,
}
\DeclareAcronym{ema}{
  short = EMA ,
  long = exponential moving average ,
}

\DeclareAcronym{rnn}{
  short = RNN ,
  long = recurrent neural network ,
}
\DeclareAcronym{lstm}{
  short = LSTM ,
  long = long short-term memory ,
}
\DeclareAcronym{gru}{
  short = GRU ,
  long = gated recurring unit ,
}
\DeclareAcronym{ssm}{
  short = SSM ,
  long = state-space model ,
}

\input{custom/abbrv.tex}

\title{Samba: Synchronized Set-of-Sequences \\Modeling for Multiple Object Tracking}

\author{
\hfill Mattia Segu$^{1,2}$, Luigi Piccinelli$^1$, Siyuan Li$^1$, Yung-Hsu Yang$^1$,  Bernt Schiele$^2$, Luc Van Gool$^{1,3}$ \hfill \\
\hfill $^1$ ETH Zurich, $^2$ Max Planck Institute for Informatics, $^3$ INSAIT \hfill  \\
\hfill \url{https://sambamotr.github.io/} \hfill 
}

\iclrfinalcopy 
\begin{document}

\maketitle

\begin{abstract}
Multiple object tracking in complex scenarios - such as coordinated dance performances, team sports, or dynamic animal groups - presents unique challenges. In these settings, objects frequently move in coordinated patterns, occlude each other, and exhibit long-term dependencies in their trajectories. However, it remains a key open research question on how to model long-range dependencies within tracklets, interdependencies among tracklets, and the associated temporal occlusions. To this end, we introduce Samba, a novel linear-time set-of-sequences model designed to jointly process multiple tracklets by synchronizing the multiple selective state-spaces used to model each tracklet. Samba autoregressively predicts the future track query for each sequence while maintaining synchronized long-term memory representations across tracklets. By integrating Samba into a tracking-by-propagation framework, we propose SambaMOTR, the first tracker effectively addressing the aforementioned issues, including long-range dependencies, tracklet interdependencies, and temporal occlusions. Additionally, we introduce an effective technique for dealing with uncertain observations (MaskObs) and an efficient training recipe to scale SambaMOTR to longer sequences. By modeling long-range dependencies and interactions among tracked objects, SambaMOTR implicitly learns to track objects accurately through occlusions without any hand-crafted heuristics. Our approach significantly surpasses prior state-of-the-art on the DanceTrack, BFT, and SportsMOT datasets. 
\end{abstract}

\section{Introduction}
\label{sec:intro}
%
%
%

\begin{figure}[t]
  \centering
  \begin{subfigure}{0.3\linewidth}
    \includegraphics[width=\linewidth, clip, trim=6.6cm 5.0cm 11.0cm 5.0cm]{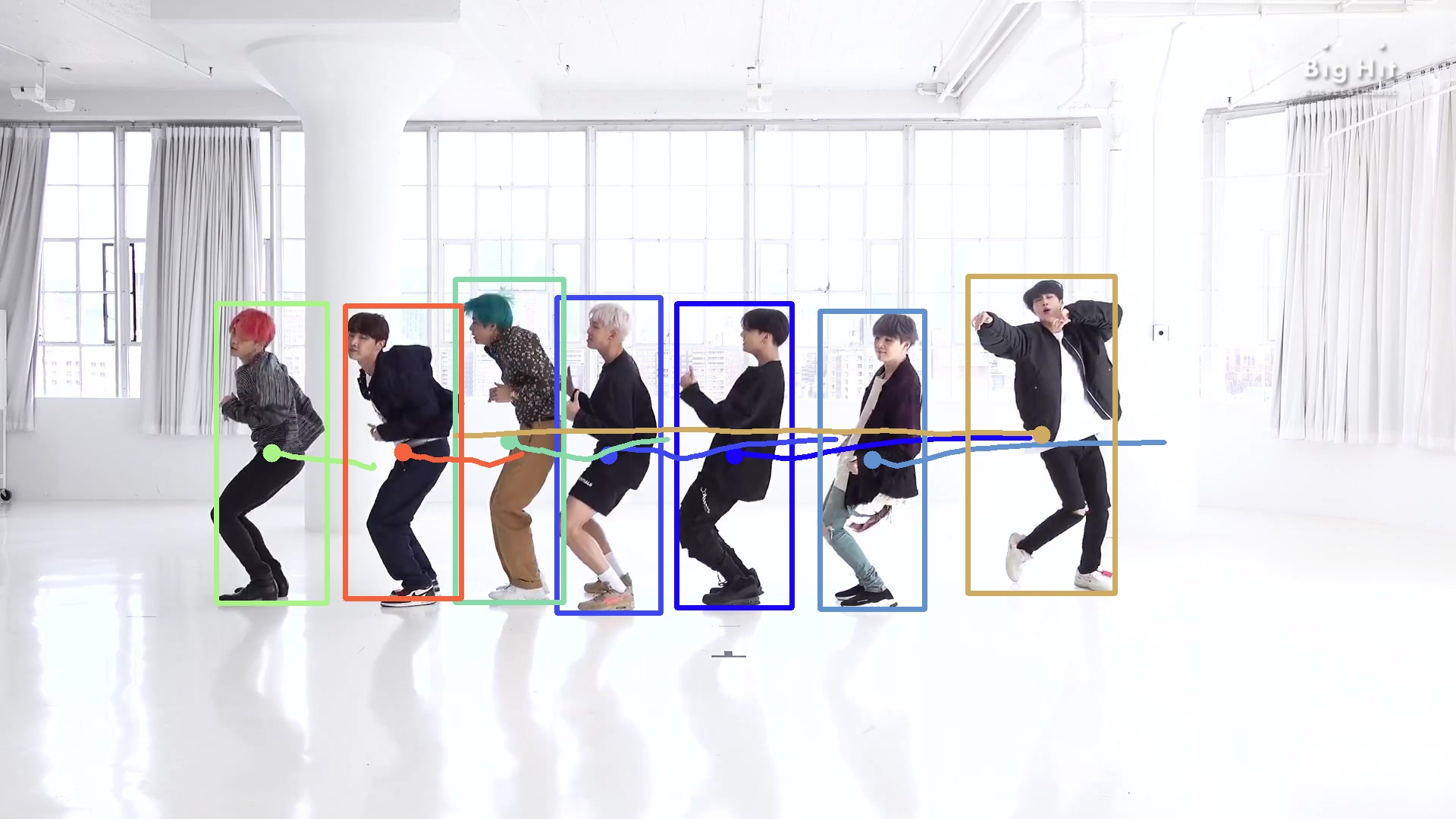}
    \caption{DanceTrack~\citep{sun2022dancetrack}}
  \end{subfigure}
  \begin{subfigure}{0.3\linewidth}
    \includegraphics[width=\linewidth, clip, trim=23cm 12.5cm 0cm 0.7cm]{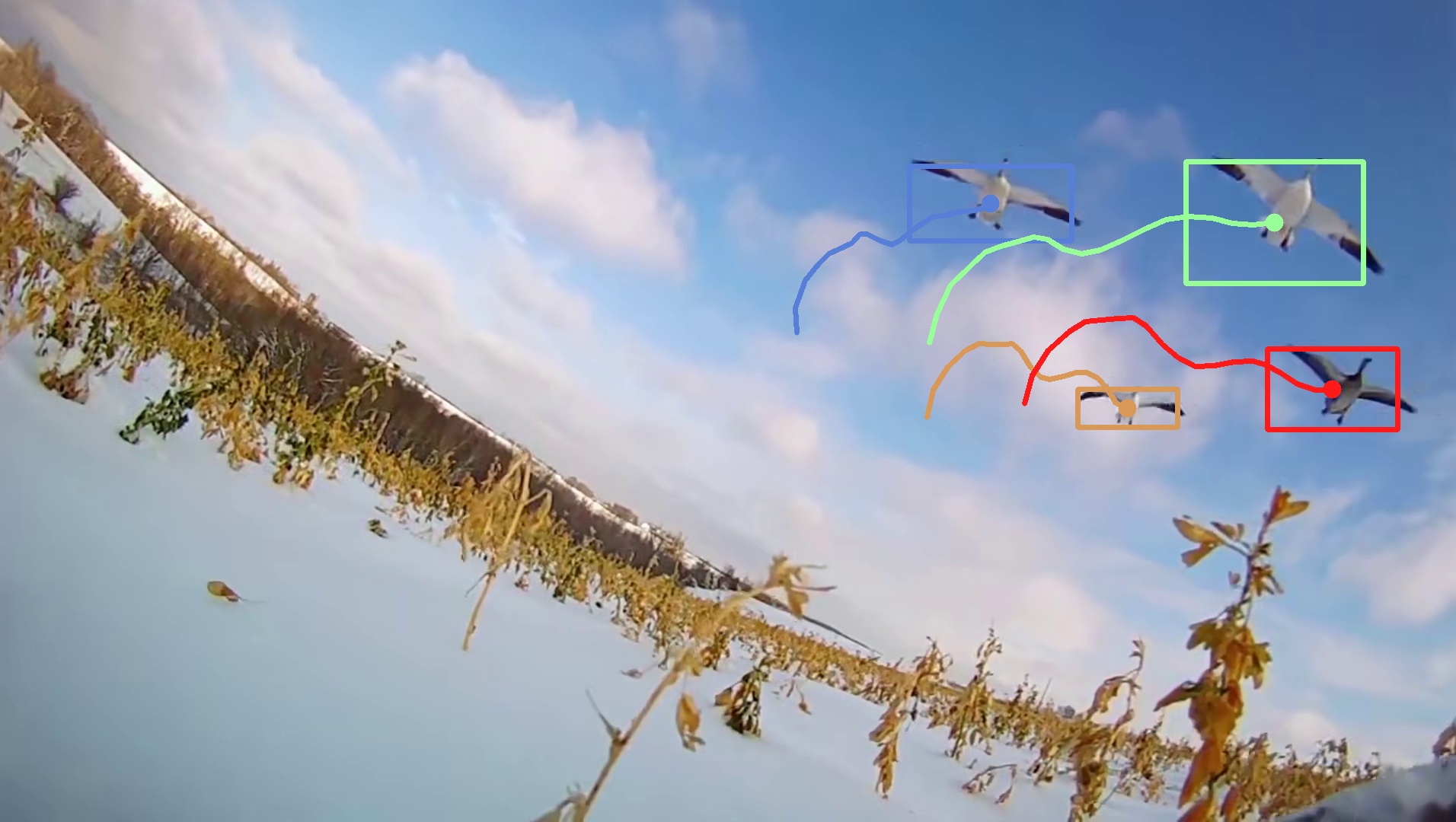}
    \caption{BFT~\citep{zheng2024nettrack}}
  \end{subfigure}
  \begin{subfigure}{0.3\linewidth}
    \includegraphics[width=\linewidth, clip, trim=3cm 4cm 11.1cm 4cm]{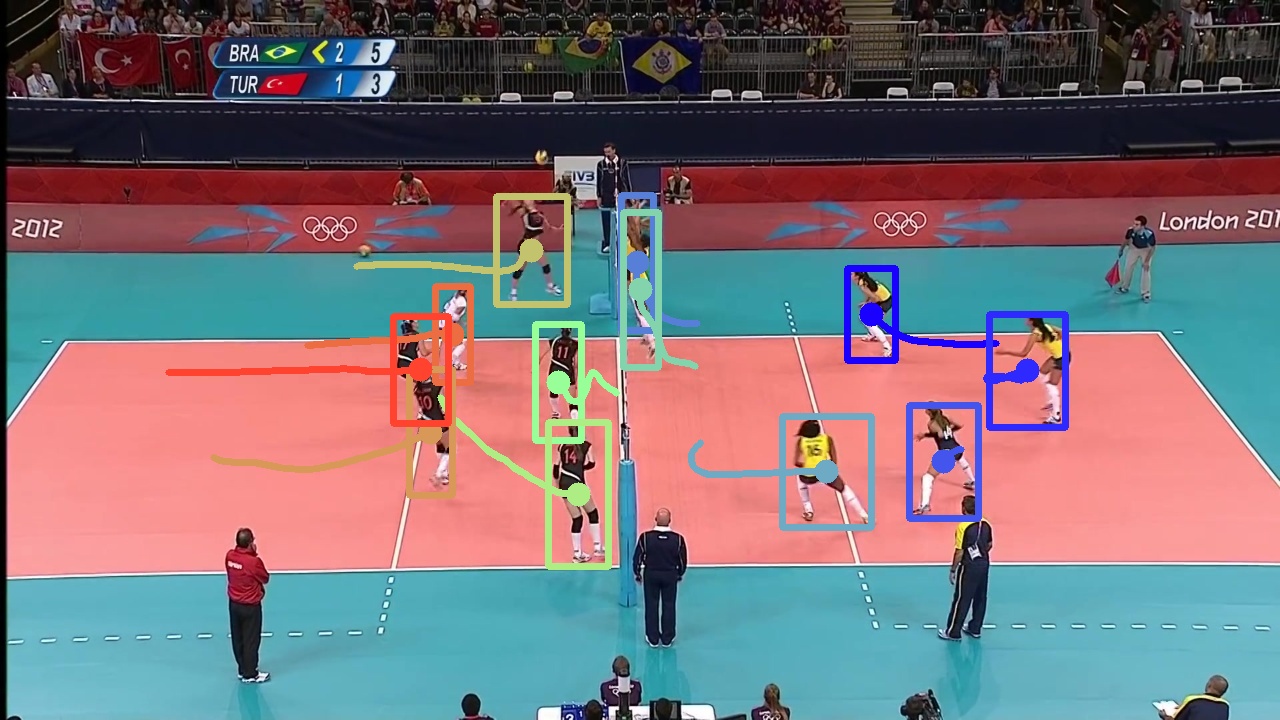}
    \caption{SportsMOT~\citep{cui2023sportsmot}}
  \end{subfigure}
  \caption{\textbf{Tracking multiple objects in challenging scenarios} - such as coordinated dance performances (a), dynamic animal groups (b), and team sports (c) - requires handling complex interactions, occlusions, and fast movements. As shown in the tracklets above, objects may move in coordinated patterns and occlude each other. By leveraging the joint long-range dependencies in their trajectories, SambaMOTR accurately tracks objects through time and occlusions.}
  \label{fig:teaser_datasets}
  \vspace{-1em}
\end{figure}

\Ac{mot} involves detecting multiple objects while keeping track of individual instances throughout a video stream. It is critical for multiple downstream tasks such as sports analysis, autonomous navigation, 
and media production~\citep{luo2021multiple}.
Traditionally, \Ac{mot} methods are validated on relatively simple settings such as surveillance datasets~\citep{milan2016mot16}, where pedestrians exhibit largely linear motion and diverse appearance, and rarely interact with each other in complex ways.
However, in dynamic environments like team sports, dance performances, or animal groups, objects frequently move in coordinated patterns, occlude each other, and exhibit non-linear motion with long-term dependencies in their trajectories (\cref{fig:teaser_datasets}).
Modeling the long-term interdependencies between objects in these settings, where their movements are often synchronized or influenced by one another, remains an open problem that current methods fail to address.

Current \textit{tracking-by-detection} methods~\citep{bewley2016simple,wojke2017simple,zhang2022bytetrack,cao2023observation} often rely on heuristics-based models like the Kalman filter to independently model the trajectory of objects and predict their future location. However, these methods struggle with the non-linear nature of object dynamics such as motion, appearance, and pose changes. 
%
\textit{Tracking-by-propagation}~\citep{sun2020transtrack,meinhardt2022trackformer,zeng2022motr} offers an alternative by modeling tracking as an end-to-end autoregressive object detection problem, leveraging detection transformers~\citep{carion2020end,zhu2020deformable} to propagate track queries over time. 
Their flexible design fostered promising performance in settings with complex motion, pose, and appearance patterns, such as dance~\citep{sun2022dancetrack}, sports~\citep{cui2023sportsmot}, and bird~\citep{zheng2024nettrack} tracking datasets. 
However, such methods only propagate the temporal information across adjacent frames, failing to account for long-range dependencies.
%
MeMOTR~\citep{gao2023memotr} attempts a preliminary solution to this problem by storing temporal information through an external heuristics-based memory. However, its use of an \ac{ema} to compress past history results in a suboptimal temporal memory representation, as it discards fine-grained long-range dependencies that are crucial for accurate tracking over time.
Moreover, by processing each tracklet independently and overlooking tracklets interaction, current methods cannot accurately model objects' behavior through occlusions, resorting to naive heuristics to handle such cases: 
some~\citep{zhang2023motrv2,gao2023memotr} freeze the track queries during occlusions and only rely on their last observed state during prolonged occlusions;
others~\citep{zeng2022motr} delegate occlusion management to the propagation module, which fails to estimate accurate track trajectories as it only propagates information across adjacent frames and does not account for historical information. 
We argue that effective long-term memory and interaction modeling allow for more accurate inference of occluded objects' behavior in complex environments, such as team sports or dance performances, by leveraging past information and understanding joint motion patterns.

To address these shortcomings, we propose Samba
\footnote{Samba is named for its foundation on the \textbf{S}ynchronization of M\textbf{amba}’s selective state-spaces~\citep{gu2023mamba}. Its name reflects the coordinated motion of tracklets (\cref{fig:teaser_datasets}), much like the synchronized movements in the samba dance. By synchronizing the hidden states across sequences, our approach is disentangled from Mamba’s selective \acp{ssm} and can, in principle, be applied to any sequence model that includes an intermediate memory representation, such as other \acp{ssm} or recurrent neural networks.}
, a novel linear-time set-of-sequences model that processes a set of sequences (\eg multiple tracklets) simultaneously and compresses their histories into synchronized long-term memory representations, capturing interdependencies within the set.
Samba adopts selective \acp{ssm} from Mamba~\citep{gu2023mamba} to independently model all tracklets, compressing their long-range histories into hidden states. 
We then propose to synchronize these memory representations across tracklets at each time step to account for interdependencies (\eg interactions among tracklets).
We implement synchronization via a self-attention mechanism~\citep{vaswani2017attention} across the hidden states of all sequences, allowing tracklets to exchange information. This approach proves beneficial in datasets where objects move in coordinated patterns (\cref{tab:dancetrack,tab:bft,tab:sportsmot}).
The resulting set-of-sequences model, Samba, retains the linear-time complexity of \acp{ssm} while modeling the joint dynamics of the set of tracklets.

By integrating Samba into a
tracking-by-propagation framework~\citep{zeng2022motr}, we present SambaMOTR, an end-to-end multiple object tracker that models long-range dependencies and interactions between tracklets to handle complex motion patterns and occlusions in a principled manner.
SambaMOTR complements a transformer-based object detector with a novel set-of-queries propagation module based on Samba, which accounts for both individual tracklet histories and their interactions when autoregressively predicting the next track queries.

\begin{figure}
  \centering
  \includegraphics[width=1.0\linewidth,trim={0.0cm 0.0cm 0.0cm 0.0cm},clip]{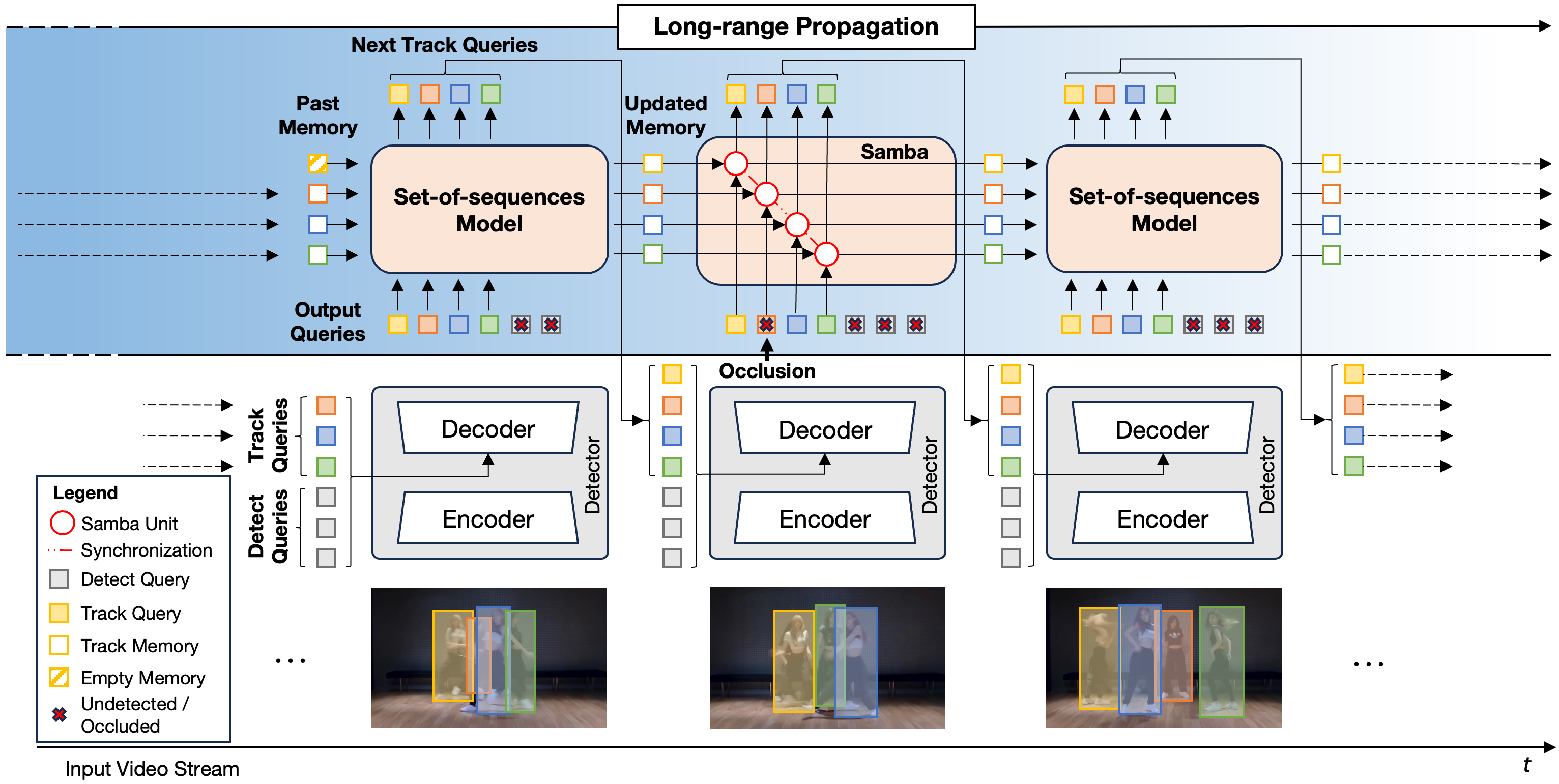}
  \caption{\textbf{Overview of SambaMOTR.} SambaMOTR combines a transformer-based object detector with a set-of-sequences Samba model. The object detector's encoder extracts image features from each frame, which are fed into its decoder together with detect and track queries to detect newborn objects or re-detect tracked ones. The Samba set-of-sequences model is composed of multiple synchronized Samba units that simultaneously process the past memory and currently observed output queries for all tracklets to predict the next track queries and update the track memory. The hidden states of newborn objects are initialized from zero values (barred squares). In case of occlusions or uncertain detections, the corresponding query is masked (red cross) during the Samba update. 
  }
  \label{fig:sambamotr}
\vspace{-1em}
\end{figure}

Additionally, some queries result in uncertain detections due to occlusions or challenging scenarios (see \cref{fig:sambamotr}, Occlusion). To prevent these detections from compromising the memory representation and accumulating errors during query propagation with Samba, we propose MaskObs. MaskObs blocks unreliable observations from entering the set-of-queries propagation module while updating the corresponding hidden states and track queries using only the long-term memory of their tracklets and interactions with confidently tracked objects. Unlike previous methods that freeze track queries during occlusions, MaskObs leverages both temporal and spatial context - \ie past behavior and interdependencies with other tracklets - to more accurately predict an object’s future state. Consequently, SambaMOTR tracks objects through occlusions more effectively (\cref{tab:method_components}, line d).

Finally, we introduce an efficient training recipe to scale SambaMOTR to longer sequences by sampling arbitrarily long sequences, computing tracking results, and applying gradients only on the last five frames. This simple strategy enables us to learn longer-range dependencies for query propagation, improving the tracking performance while maintaining the same GPU memory requirements as previous methods~\citep{zeng2022motr,gao2023memotr}.

We validate SambaMOTR on the challenging DanceTrack~\citep{sun2022dancetrack}, SportsMOT~\citep{cui2023sportsmot}, and BFT~\citep{zheng2024nettrack} datasets. Owing to our contributions, we establish a new state of the art on all datasets. We summarize them as follows:
\textbf{(a)} we introduce Samba, our novel linear-time set-of-sequences model based on synchronized \acp{ssm};
\textbf{(b)} we introduce SambaMOTR, the first tracking-by-propagation method that leverages past tracklet history in a principled manner to learn long-range dependencies, tracklets interaction, and occlusion handling; 
\textbf{(c)} we introduce MaskObs, a simple technique for dealing with uncertain observations in \acp{ssm} and an efficient training recipe that enables learning stronger sequence models with limited compute.

\vspace{-0.5em}
\section{Related Work}
\label{sec:related}
\vspace{-0.5em}
\paragraph{Tracking-by-detection.} \textit{Tracking-by-detection} is a popular paradigm in \ac{mot}, consisting of an object detection stage followed by data association to yield object trajectories throughout a video.
Motion and appearance cues are typically utilized to match detections to tracklets through hand-crafted heuristics.
The motion-based tracker SORT~\citep{bewley2016simple} relies on \ac{iou} to assign the tracklet locations predicted with a Kalman filter to object detections. 
ByteTrack~\citep{zhang2022bytetrack} introduces a two-stage matching scheme to associate low-confidence detections.
OC-SORT~\citep{cao2023observation} models non-linear motion by taking care of noise accumulation under occlusion.
Alternatively, appearance descriptors can be used alone~\citep{pang2021quasi,li2022tracking, MASA} or in combination with motion~\citep{wojke2017simple,wang2020towards,zhang2021fairmot,segu2024walker,li2024slack} to 
match detections to tracklets according to a similarity metric. 
Due to the disentangled nature of the two stages, tracking-by-detection methods historically leveraged state-of-the-art object detectors to top the MOT challenge~\citep{dendorfer2021motchallenge}. 
However, by relying on hand-crafted heuristics, such methods struggle with non-linear motion and appearance patterns~\citep{sun2022dancetrack,cui2023sportsmot,zheng2024nettrack,li2023ovtrack}, and require domain-specific hyperparameters~\citep{segu2023darth,liu2023cooler}. 
Recent transformer-based methods have eased the burden of heuristics.  
TransTrack~\citep{sun2020transtrack} decodes track and detect queries with siamese transformer decoders and associates them with simple \ac{iou} matching.
MeMOT~\citep{cai2022memot} fuses a large memory bank into a tracklet descriptor with a transformer-based memory aggregator. However, it requires storing and processing with quadratic complexity the historical information from up to 27 past frames.
GTR~\citep{zhou2022global} matches static trajectory queries to detections to generate tracklets, but fails to model object motion and tracklet interaction.
In contrast, SambaMOTR implicitly learns motion, appearance, and tracklet interaction models by autoregressively predicting the future track queries with our set-of-sequences model Samba.

\paragraph{Tracking-by-propagation.}
Recent work~\citep{meinhardt2022trackformer,zeng2022motr} introduced a more flexible and end-to-end trainable \emph{tracking-by-propagation} design that treats \ac{mot} as an autoregressive problem where object detection and query propagation are tightly coupled. 
Leveraging the transformer-based Deformable DETR~\citep{zhu2020deformable} object detector, TrackFormer~\citep{meinhardt2022trackformer} and MOTR~\citep{zeng2022motr} autoregressively propagate the detection queries through time to re-detect (track) the same object in subsequent frames.
MOTRv2~\citep{zhang2023motrv2} leverages a pre-trained YOLOX object detector to provide anchors for Deformable DETR and boost its detection performance.
However, these approaches only propagate queries across adjacent frames, failing to fully leverage the historical information. 
MeMOTR~\citep{gao2023memotr} first attempts to utilize the temporal information in tracking-by-propagation by aggregating long- (\ac{ema} of a tracklet's queries through time) and short-term memory (fusion of the output detect queries across the last two observed frames) in a temporal interaction module.
By collapsing the tracklet history with an \ac{ema} and by freezing the last observed state of track queries and memory under occlusions, MeMOTR cannot accurately estimate track query trajectories through occlusions.
Finally, by modeling query propagation independently for each tracklet, it does not model tracklet interaction.
In contrast, our proposed Samba set-of-sequences model relies on individual \acp{ssm} to independently model each tracklet as a sequence and it synchronizes the memory representations across the set of tracklets to enable tracklet interaction. Equipped with Samba, SambaMOTR autoregressively predicts future queries aware of long-range dynamics and of other tracklets' motion and appearance.

\vspace{-0.5em}
\section{Preliminaries} \label{ssec:preliminaries}
\vspace{-0.5em}

Before introducing SambaMOTR (\cref{sec:method}), we present the necessary background and notation on selective state-space models (\cref{ssec:selective}) and tracking-by-propagation (\cref{ssec:tracking-by-propagation}).

\subsection{Selective State-Space Models} \label{ssec:selective}
Inspired by classical state-space models (SSMs)~\citep{kalman1960new}, structured SSMs (S4)~\citep{gu2021efficiently} introduce a sequence model whose computational complexity scales linearly, rather than quadratically, with the sequence length. This makes S4 a principled and efficient alternative to transformers~\citep{vaswani2017attention}.
By further introducing a selection mechanism - \ie rendering the \ac{ssm} parameters input-dependent - Mamba~\citep{gu2023mamba} can model time-variant systems, bridging the performance gap with transformers~\citep{vaswani2017attention}.

We here formally define selective \acp{ssm} (S6)~\citep{gu2023mamba}. 
Let $x(t)$ be the input signal at time $t$, $h(t)$ the hidden state, and $y(t)$ the output signal.
Given the system $\mathbf{A}$, control $\mathbf{B}$, and output $\mathbf{C}$ matrices, we define the continuous linear time-variant \ac{ssm} in \cref{eq:continuous}.
The discrete-time equivalent system (\cref{eq:discrete}) of the defined \ac{ssm}  is obtained through a discretization rule.
The chosen discretization rule is typically the zero-order hold (ZOH) model: $\mathbf{\bar{A}}(t) = \exp(\mathbf{\Delta}(t) \mathbf{A}(t))$, \quad $\mathbf{\bar{B}}(t) = (\mathbf{\Delta}(t) \mathbf{A}(t))^{-1}(\exp(\mathbf{\Delta}(t) \mathbf{A}(t)) - \mathbf{I}) \cdot \mathbf{\Delta}(t) \mathbf{B}(t)$:
\noindent
\begin{minipage}{.5\linewidth}
\small
\begin{align} 
h'(t) &= \mathbf{A}(t)h(t) + \mathbf{B}(t)x(t)\label{eq:continuous} \\
y(t) &= \mathbf{C}(t)h(t) \notag
\end{align} 
\end{minipage}%
\begin{minipage}{.5\linewidth}
\small
\begin{align} 
h_t &= \mathbf{\bar{A}}(t)h_{t-1} + \mathbf{\bar{B}}(t)x_t \label{eq:discrete} \\
y_t &= \mathbf{C}(t)h_t \notag
\end{align} 
\end{minipage}

$x_t$, $h_t$, $y_t$ are the observations sampled at time $t$ of the input signal $x(t)$, hidden state $h(t)$, and output signal $y(t)$.
While S4 learns a linear time-invariant (LTI) system with $\mathbf{\Delta}(t) \!=\! \mathbf{\Delta}$, $\mathbf{A}(t) \!=\! \mathbf{A}$, $\mathbf{B}(t) \!=\! \mathbf{B}$ and $\mathbf{C}(t) \!=\! \mathbf{C}$, S6 introduces selectivity to learn a time-variant system by making $\mathbf{\Delta}(t)$, $\mathbf{B}(t)$ and $\mathbf{C}(t)$ dependent on the input $x(t)$, \ie $\mathbf{\Delta}(t) = \tau_{\Delta}(\mathbf{\Delta} + s_{\Delta}(x(t)))$, $\mathbf{B}(t) = s_B(x(t))$, $\mathbf{C}(t) = s_C(x(t))$, where $\tau_{\Delta} = \texttt{softplus}$, and $s_{\Delta}$, $s_B$, $s_C$ are learnable linear mappings.

In this paper, we propose to treat tracking-by-propagation as a sequence modeling problem. Given the discrete sequence of historical track queries for a certain tracklet, our query propagation module Samba (\cref{ssec:samba}) leverages \acp{ssm} to account for the historical tracklet information in a principled manner. By recursively compressing all tracklet history into a long-term memory, Samba's complexity scales linearly with the number of frames, enabling efficient training on long sequences while processing indefinitely long tracklets at inference time.
%
\subsection{Tracking-by-propagation} \label{ssec:tracking-by-propagation}
Tracking-by-propagation methods alternate between a detection stage and a propagation stage, 
relying on a DETR-like~\citep{carion2020end} transformer object detector and a query propagation module. 
At a time step $t$, the backbone and transformer encoder extract image features for a frame $I_t$.
The \textit{detection stage} involves feeding both a fixed-length set of learnable detect queries $Q^{det}_t$ to the transformer decoder to detect newborn objects and a variable-length set of propagated track queries $Q^{tck}_t$ to re-detect tracked ones.
At time $t=0$, the set of track queries is empty, \ie $Q^{tck}_0 = E^{tck}_0 = \emptyset$.
Detect and track queries $[Q^{det}_t, Q^{tck}_t]$ interact in the decoder with image features to generate the corresponding output embeddings $[E^{det}_t, E^{tck}_t]$ and bounding box predictions $[D^{det}_t, D^{tck}_t]$.
We denote the set of embeddings corresponding to newborn objects as $\hat{E}^{det}_t$, and $\hat{E}^{tck}_t = [\hat{E}^{det}_t, E^{tck}_t]$ as the set of embeddings corresponding to the tracklets $\mathcal{S}_t$ active at time $t$.
During the \textit{propagation stage}, a query propagation module $\Theta(\cdot)$ typically takes as input the set of embeddings $\hat{E}^{tck}_t$ and outputs refined tracked queries $Q^{tck}_{t+1} = \Theta(\hat{E}^{tck}_t)$ to re-detect the corresponding objects in the next frame.
%

Although prior work failed to properly model long-range history and tracklet interactions~\citep{zeng2022motr,gao2023memotr,meinhardt2022trackformer}, and given that multiple objects often move synchronously (\cref{fig:teaser_datasets}), we argue that the future state of objects in a scene can be better predicted by (i) considering both their historical positions and appearances, and (ii) estimating their interactions.
In this work, we cast query propagation as a set-of-sequences modeling problem.
Given a set of multiple tracklets, we encode the history of each tracklet in a memory representation using a state-space model and propose memory synchronization to account for their joint dynamics.
\vspace{-0.5em}
\section{Method} \label{sec:method}
\vspace{-0.5em}
In this section, we introduce SambaMOTR, an end-to-end multiple object tracker that combines transformer-based object detection with our set-of-sequences model Samba to jointly model the long-range history of each tracklet and the interaction across tracklets to propagate queries.
First, in \cref{ssec:tracking-by-propagation} we provide background on the tracking-by-propagation framework and motivate the need for better modeling of both temporal information and tracklets interaction.
Then, we describe the SambaMOTR architecture (\cref{ssec:sambamotr_architecture}) and introduce Samba (\cref{ssec:samba}), our novel set-of-sequences model based on synchronized state spaces that jointly models the temporal dynamics of a set of sequences and their interdependencies.
Finally, in \cref{ssec:sambamotr_propagation} we describe SambaMOTR's query propagation strategy based on Samba, our effective technique MaskObs to deal with occlusions in \acp{ssm}, a recipe to learn long-range sequence models with limited compute, and our simple inference pipeline.


\vspace{-0.5em}
\subsection{Architecture} \label{ssec:sambamotr_architecture}
\vspace{-0.5em}
Similar to other tracking-by-propagation methods~\citep{meinhardt2022trackformer,zeng2022motr,gao2023memotr}, the proposed SambaMOTR architecture (\cref{fig:sambamotr}) is composed of a DETR-like~\citep{carion2020end} object detector and a query propagation module.
As object detector, we use Deformable-DETR~\citep{zhu2020deformable} with a ResNet-50~\citep{he2016deep} backbone followed by a transformer encoder to extract image features and a transformer decoder to detect bounding boxes from a set of detect and track queries.
As query propagation module, we use our set-of-sequences model Samba. Each sequence is processed by a Samba unit synchronized with all others. A Samba unit consists of two Samba blocks (\cref{ssec:samba}) interleaved with LayerNorm~\citep{ba2016layer} and a residual connection.  
%

\subsection{Samba: Synchronized State-Space Models for Set-of-sequences Modeling} \label{ssec:samba}

\begin{figure}

  \centering
  \includegraphics[width=0.95\linewidth,trim={0.0cm 0.0cm 0.0cm 0.0cm},clip]{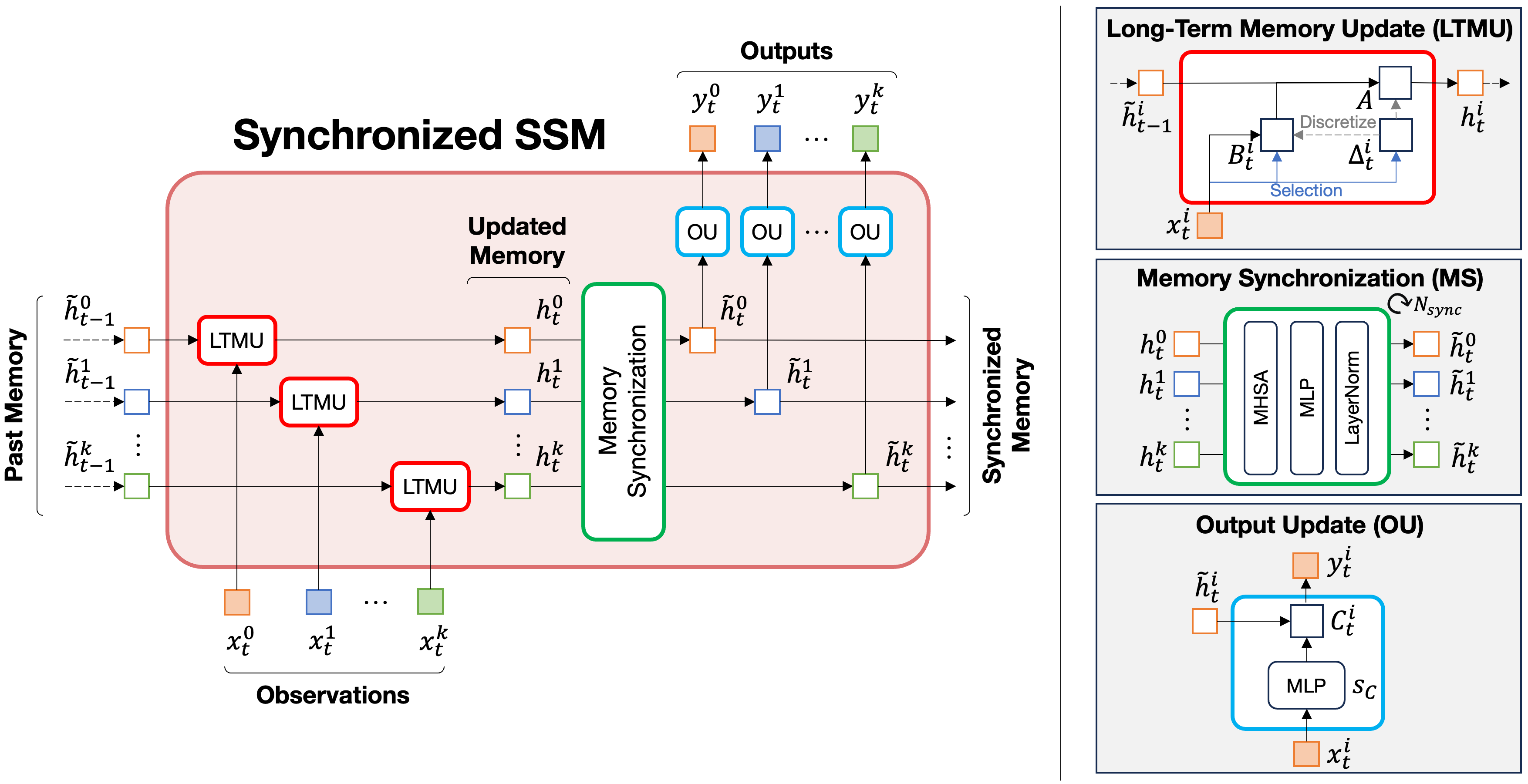}
  \caption{\textbf{Synchronized State-Space Models.} We illustrate a set of $k$ synchronized \acp{ssm}. A Long-Term Memory Update block updates each hidden state $\Tilde{h}_{t-1}^i$ based on the current observation $x_t^i$, resulting in the updated memory $h_t^i$. The Memory Synchronization block then derives the synchronized hidden state $\Tilde{h}_t^i$, which is fed into the Output Update module to predict the output $y_t^i$.}
  \label{fig:ssm}
\vspace{-1em}
\end{figure}
Set-of-sequences modeling involves simultaneously modeling a set of temporal sequences and the interdependencies among them. In \ac{mot}, set-of-sequences models can capture long-range temporal relationships within each tracklet as well as complex interactions across tracklets. 
To this end, we introduce Samba, a linear-time set-of-sequences model based on the synchronization of multiple state-space models.
In this paper, we leverage Samba as a set-of-queries propagation network to jointly model multiple tracklets and their interactions in a tracking-by-propagation framework.

\myparagraph{Synchronized Selective State-Space Models.} Let $x_t^i$ be the discrete observation at time $t$ of the $i$-th input sequence from a set of sequences $\mathcal{S}$. We choose selective \acp{ssm}~\citep{gu2023mamba} to model each sequence through a hidden state $h_t^i$ (\cref{eq:samba_s6}) encoding long-term memory, but our approach applies to any other \ac{ssm}. 
Given the memory $\Tilde{h}_{t-1}^i$, we define a long-term memory update (LTMU) function that updates $\Tilde{h}_{t-1}^i$ based on the current observation $x_t^i$, resulting in the updated memory $h_t^i$.
We propose a memory synchronization (MS) function $\Gamma_{i \in \mathcal{S}}(\cdot)$ that produces a set of synchronized hidden states $\Tilde{h}_t^i$ $\forall i \in \mathcal{S}$, modeling interactions across the set of sequences $\mathcal{S}$ (\cref{eq:samba_sync}). 
Finally, we derive the output $y_t^i$ for each sequence through the output update (OU) function (\cref{eq:samba_output}).
\begin{subequations} \label{eq:samba}
\begin{align} 
\text{LTMU}  \! : \qquad h_t^i &= \mathbf{\bar{A}}^i(t)h_{t-1}^i + \mathbf{\bar{B}}^i(t)x_t^i \label{eq:samba_s6} \\ 
\text{MS}   \! : \qquad \Tilde{h}_t^i &= \Gamma_{i \in \mathcal{S}}\left(h_t^i\right) \label{eq:samba_sync} \\ 
\text{OU}  \! : \qquad y^i_t &= \mathbf{C}(t)\Tilde{h}^i_t \label{eq:samba_output}
\end{align}
\end{subequations}
An ideal memory synchronization function should be flexible regarding the number of its inputs (hidden states) and equivariant to their order.
Thus, we propose to define the memory synchronization function $\Gamma(\cdot) = [FFN(MHSA( \cdot ))]_{\times N_{sync}}$ as a set of $N_{sync}$ stacked blocks with multi-head self-attention (MHSA)~\citep{vaswani2017attention} followed by a feed-forward network (FFN).
A schematic illustration of the proposed synchronized state-space model layer is in \cref{fig:ssm}. 

\myparagraph{Set-of-sequences Model.} Refer to \cref{sec:app_samba} for a detailed description of how our synchronized SSM is used in the Samba units that model each sequence in the set-of-sequences Samba model.


\subsection{SambaMOTR: End-to-end Tracking-by-propagation with Samba}
\label{ssec:sambamotr_propagation}
\myparagraph{Query Propagation with Samba.}
As described in \cref{ssec:tracking-by-propagation}, the query propagation module $\Theta(\cdot)$ takes as input the decoder output embeddings $\hat{E}^{tck}_t$ and outputs refined track queries $Q^{tck}_{t+1} = \Theta(\hat{E}^{tck}_t)$.
SambaMOTR extends this paradigm by accounting for the temporal information and tracklets interaction.
In particular, we use our Samba module $\Phi(\cdot)$ to compress the history of each tracklet into a hidden state $h^i_{t}$ and synchronize it across tracklets to derive the synchronized memory $\Tilde{h}^i_{t}$. 
Notice that $\Tilde{h}^i_t=\mathbf{0}$ for a newborn object $i$.
At time $t$, we first enrich the detector output embeddings $\hat{E}^{tck}_t$ with position information by summing to them sine-cosine positional encodings $PE(\cdot)$ of the corresponding bounding boxes coordinates $\hat{D}^{tck}_t$ to implicitly model object motion and appearance, obtaining the set of input observations  $X^{tck}_t = \hat{E}^{tck}_t + PE(\hat{D}^{tck}_t)$.
Given the set of input observations $X^{tck}_t$ and past synchronized hidden states $\Tilde{H}_{t-1}$ for all tracklets in the set $\mathcal{S}_t$ of tracklets at time $t$, we feed them into Samba $(Y_t, \Tilde{H}_t) = \Phi(X^{tck}_t, \Tilde{H}_{t-1})$ to obtain the output embeddings $Y_t$ and updated synchronized hidden states $\Tilde{H}_t$.
Finally, we use the output embeddings $Y_t$ and a learnable mapping $s_y$ to predict a residual $\Delta Q^{tck}_t = s_y(Y_t)$ to the past track queries $Q^{tck}_t$ and generate the new ones, \ie $Q^{tck}_{t+1} = Q^{tck}_t + \Delta Q^{tck}_t$.
By recursively unfolding this process over time, SambaMOTR can track multiple objects while compressing indefinitely long tracklet histories into their long-term memory representations, effectively modeling object motion and appearance changes, and tracklets interactions.

\myparagraph{MaskObs: Dealing with Uncertain Observations.}
Tracking-by-propagation may occasionally deal with occluded objects or uncertain detections.
Given a function $conf(\cdot)$ to estimate the predictive confidence of an input observation $x^i_t$, we propose MaskObs, a strategy to handle uncertain observations.
MaskObs masks uncertain observations from the state update (\cref{eq:occlusion_masking}), thus defining the system dynamics solely based on its history and the interdependencies with other sequences:
\begin{align}
\label{eq:occlusion_masking}
\vspace{-30pt}
h_t^i &= \mathbf{\bar{A}^i}(t)h^i_{t-1} + \mathbf{\bar{B}^i}(t)x^i_t \cdot \mathds{1}[conf(x^i_t) > \tau_{mask}] 
\vspace{-25pt}
\end{align}
$\mathds{1}[\cdot]$ is the indicator function, and $\tau_{mask}$ is the confidence threshold, \eg $\tau_{mask}=0.5$.
We implement $conf(x^i_t)$ as the predictive confidence $conf(d^i_t)$ of the corresponding bounding box $d^i_t$. 
In our work, this design choice allows us to better model query propagation through occlusions (\cref{tab:method_components}, line b).

\myparagraph{Efficiently Learning Long-range Sequence models.}
Previous MOTR-like approaches are trained end-to-end on a sequence of $5$ consecutive frames sampled at random intervals. 
While SambaMOTR's set-of-sequences model Samba already shows impressive generalization performance to long sequences at inference time (\cref{tab:method_components}, line c), we propose to train on longer sequences (\ie $10$ frames) and only apply gradients to the last $5$ frames (\cref{tab:method_components}, line d). 
We hypothesize that this strategy allows us to learn better history compression for late observations in a sequence, resulting in even better tracking performance while being trained with similar GPU memory requirements.
A schematic illustration of our training scheme proposal is in \cref{fig:method_components}.
 
\myparagraph{Inference Pipeline.}
At a given time step $t$, we jointly input the learnable detect queries $Q_{det}$ and track queries $Q_{t}^{tck}$ ($Q_{0}^{tck} = \emptyset$) into the transformer decoder to produce detection embeddings $E^{det}_t$ and tracking embeddings $E_t^{tck}$ and the corresponding bounding boxes.
Each detection bounding box with a confidence score higher than a threshold $\tau_{det}$ will initialize a newborn track $\hat{E}^{det}$.
We then propagate the embeddings of newborn $\hat{E}^{det}$ and tracked $E_t^{tck}$ objects together with the track memory $\Tilde{H}_{t-1}$ to generate the updated track queries $Q^{t+1}_{track}$ and synchronized memory $\Tilde{H}_{t+1}$.
To deal with occlusions and lost objects, we consider an individual track query $q_t^{i, track}$ inactive if its corresponding bounding box confidence $conf(d^i_t)$ at time $t$ is lower than $\tau_{track}$. If a track query is inactive for more than $N_{miss}$ frames, it is deemed lost and dropped.

Unlike MeMOTR~\citep{gao2023memotr}, which does not update the track embedding and long-term memory for an object with low detection confidence at a time step $t$,  our approach employs a principled query propagation scheme that can hallucinate likely track query trajectories under occlusions by relying on its past history or attending to other trajectories.
Thus, we always update the memory and track query for any tracklet - even when occluded - as long as it is not deemed lost.
\vspace{-0.5em}
\section{Experiments}
\vspace{-0.5em}
In this section, we present experimental results to validate SambaMOTR.
We describe our evaluation protocol (\cref{ssec:eval_protocol}) and report implementation details (\cref{ssec:implementation}).
We then compare SambaMOTR to the previous state-of-the-art methods (\cref{ssec:exp_sota}) and conduct an ablation study (\cref{ssec:exp_ablation}) on the method components.
We provide more ablations in the appendix.
Qualitative results can be found in \cref{fig:teaser_datasets} and at the anonymous project page \url{https://anonymous-samba.github.io/}.

\subsection{Evaluation Protocol} \label{ssec:eval_protocol}
\paragraph{Datasets.} To evaluate SambaMOTR, we select a variety of challenging datasets exhibiting highly non-linear motion in crowded scenarios, with frequent occlusions and uniform appearances.
All datasets present scenes with objects moving synchronously. Thus, they represent a suitable benchmark for assessing the importance of modeling tracklet interaction.
DanceTrack~\citep{sun2022dancetrack} is a multi-human tracking dataset composed of 100 group dancing videos. 
The Bird Flock Tracking (BFT)~\citep{zheng2024nettrack} dataset includes 106 clips from the BBC documentary series Earthflight~\citep{downer2011earthflight}. 
SportsMOT~\citep{cui2023sportsmot} consists of 240 video sequences from basketball, volleyball, and soccer scenes. 
Due to the highly linear motion in MOT17~\citep{milan2016mot16}, its small size (only 7 videos), and the subsequent need for training on additional detection datasets, end-to-end tracking methods do not provide additional advantages over more naive Kalman-filter-based methods. We report its results in the Appendix. 

\paragraph{Metrics.} Following prior work, we measure the overall tracking performance with the HOTA~\citep{luiten2021hota} metric and disentangle detection accuracy (DetA) and association accuracy (AssA).
We report the MOTA~\citep{bernardin2008evaluating} and IDF1~\citep{ristani2016performance} metrics for completeness. 
Since our objective is improving association performance and the overall tracking quality, HOTA and AssA are the most representative metrics.

\subsection{Implementation Details} \label{ssec:implementation}
Following prior works~\citep{gao2023memotr,zhang2023motrv2}, we apply random resize, random crop, and photometric augmentations as data augmentation.
The shorter side of the input image is resized to 800 preserving the aspect ratio, and the maximum size is restricted to 1536.
For a fair comparison with prior work~\citep{sun2020transtrack, zeng2022motr, gao2023memotr}, we use the Deformable-DETR~\citep{zhu2020deformable} object detector with ResNet-50~\citep{he2016deep} and initialize it from COCO~\citep{lin2014microsoft} pre-trained weights.
Similar to MeMOTR~\citep{gao2023memotr}, we inject track queries after one decoder layer.
We run our experiments on 8 NVIDIA RTX 4090 GPUs, with batch size 1 per GPU.
Each batch element contains a video clip with 10 frames, and we compute and backpropagate the gradients only over the last 5.
We sample uniformly spaced frames at random intervals from 1 to 10 within each clip.
We utilize the AdamW optimizer~\citep{loshchilov2017decoupled} with initial learning rate of $2.0 \times 10^{-4}$.
For simplicity, $\tau_{det}\!=\!\tau_{track}\! = \!\tau_{mask} \!= \!0.5$. $N_{miss}$ is $35$, $20$, and $50$ on DanceTrack, BFT, and SportsMOT, respectively, due to different dataset dynamics. 
On DanceTrack~\citep{sun2022dancetrack}, we train SambaMOTR for 15 epochs on the training set and drop the learning rate by a factor of $10$ at the $10^{th}$ epoch.
On BFT~\citep{sun2022dancetrack}, we train for 20 epochs and drop the learning rate after 10 epochs.
On SportsMOT~\citep{cui2023sportsmot}, we train for 18 epochs and drop the learning rate after 8 and 12 epochs.
SambaMOTR's inference runs at 16 FPS on a single NVIDIA RTX 4090 GPUs.

\subsection{Comparison with the State of the Art} \label{ssec:exp_sota}
We compare SambaMOTR with multiple tracking-by-detection and tracking-by-propagation approaches on the DanceTrack (\cref{tab:dancetrack}), BFT (\cref{tab:bft}) and SportsMOT (\cref{tab:sportsmot}) datasets. 
All methods are trained without using additional datasets.
Since trackers use various object detectors with different baseline performance, we report the detector used for each method.
For fair comparison, we report the performance of tracking-by-propagation methods with Deformable DETR~\citep{zhu2020deformable}, marking the best in \textbf{bold}. We underle the overall best result. Tracking-by-detection methods often use the stronger YOLOX-X~\citep{ge2021yolox}, but tracking-by-propagation consistently outperforms them, with SambaMOTR achieving the highest HOTA and AssA across all datasets.

\begin{table}[t]
  \caption{\textbf{State-of-the-art comparison on DanceTrack}~\citep{sun2022dancetrack} without additional training data. Best tracking-by-propagation method in \textbf{bold}; best overall \underline{underlined}.}
  \label{tab:dancetrack}
  \centering
\tablesize
  \setlength{\tabcolsep}{5pt}{
    \begin{tabular}{llccccc}
      \toprule
      Methods & Detector & HOTA & AssA & DetA & IDF1 & MOTA \\
      \midrule
      \bf \textit{Tracking-by-detection:} \\
      FairMOT~\citep{zhang2021fairmot} & \multirow{3}{*}{CenterNet~\citep{duan2019centernet}} & 39.7 & 23.8 & 66.7 & 40.8 & 82.2 \\
      CenterTrack~\citep{zhou2020tracking} &  & 41.8 & 22.6 & 78.1 & 35.7 & 86.8 \\
      TraDeS~\citep{wu2021track} &  & 43.3 & 25.4 & 74.5 & 41.2 & 86.2 \\
      \greyrule
      TransTrack~\citep{sun2020transtrack} & Deformable DETR~\citep{zhu2020deformable} & 45.5 & 27.5 & 75.9 & 45.2 & 88.4 \\
      \greyrule
      GTR~\citep{zhou2022global} & CenterNet2~\citep{zhou2021probabilistic} & 48.0 & 31.9 & 72.5 & 50.3 & 84.7 \\
      \greyrule
      ByteTrack~\citep{zhang2022bytetrack} & \multirow{4}{*}{YOLOX-X~\citep{ge2021yolox}} & 47.7 & 32.1 & 71.0 & 53.9 & 89.6 \\
      QDTrack~\citep{pang2021quasi} &  & 54.2 & 36.8 & 80.1 & 50.4 & 87.7 \\
      OC-SORT~\citep{cao2023observation} &  & 55.1 & 38.3 & 80.3 & 54.6 & \underline{92.0} \\
      C-BIoU~\citep{yang2023hard} &  & 60.6 & 45.4 & \underline{81.3} & 61.6 & 91.6 \\
      \midrule
      \bf \textit{Tracking-by-propagation:} \\
      MOTR~\citep{zeng2022motr} & \multirow{3}{*}{Deformable DETR~\citep{zhu2020deformable}} & 54.2 & 40.2 & 73.5 & 51.5 & 79.7 \\
      MeMOTR~\citep{gao2023memotr} & & 63.4 & 52.3 & 77.0 & 65.5 & 85.4 \\
      SambaMOTR (ours) & & \bf \underline{67.2} & \bf \underline{57.5} & \bf 78.8 & \bf \underline{70.5} & \bf 88.1 \\
      \bottomrule
    \end{tabular}
  }
\end{table}

\begin{table}[ht]
\caption{\textbf{State-of-the-art comparison on BFT}~\citep{zheng2024nettrack} without additional training data. Best tracking-by-propagation method in \textbf{bold}; best overall \underline{underlined}.}
\label{tab:bft}
\centering
\tablesize
\setlength{\tabcolsep}{5pt}{
\begin{tabular}{@{}llccccc@{}}
\toprule
Method           & Detector                         & HOTA          & AssA          & DetA          & IDF1          & MOTA          \\ \midrule
    \bf \textit{Tracking-by-detection:} \\
    FairMOT~\citep{zhang2021fairmot}          & \multirow{2}{*}{CenterNet~\citep{duan2019centernet}}       & 40.2          & 28.2          & 53.3          & 41.8          & 56.0          \\
    CenterTrack~\citep{zhou2020tracking}      &                                  & 65.0          & 54.0          & 58.5          & 61.0          & 60.2          \\ \greyrule
    SORT~\citep{wojke2017simple}             & \multirow{3}{*}{YOLOX-X~\citep{ge2021yolox}}           & 61.2          & 62.3          & 60.6          & 77.2          & 75.5          \\
    ByteTrack~\citep{zhang2022bytetrack}        &                                  & 62.5          & 64.1          & 61.2          & \underline{82.3} & \underline{77.2} \\
    OC-SORT~\citep{cao2023observation}          &                                  & 66.8          & 68.7          & 65.4          & 79.3          & 77.1          \\ \greyrule
    TransCenter~\citep{xu2022transcenter}      & \multirow{2}{*}{Deformable DETR~\citep{zhu2020deformable}} & 60.0          & 61.1          & \underline{66.0} & 72.4          & 74.1          \\
    TransTrack~\citep{sun2020transtrack}       &                                  & 62.1          & 60.3          & 64.2          & 71.4          & 71.4          \\
    \midrule
    \bf \textit{Tracking-by-propagation:} \\
    TrackFormer~\citep{meinhardt2022trackformer}      &    \multirow{2}{*}{Deformable DETR~\citep{zhu2020deformable}}  & 63.3          & 61.1          & \textbf{\underline{66.0}} & 72.4          & \bf 74.1          \\
    SambaMOTR (ours) &                                  & \textbf{\underline{69.6}} & \textbf{\underline{73.6}} & \textbf{\underline{66.0}} & \bf 81.9          & 72.0          \\ \bottomrule
\end{tabular}}
\end{table}
\paragraph{DanceTrack.} The combination of highly irregular motion and crowded scenes with frequent occlusions and uniform appearance historically made DanceTrack challenging for tracking-by-detection methods. Despite their higher DetA when using the strong object detector YOLOX-X~\citep{ge2021yolox}, tracking-by-propagation significantly outperforms them (see MeMOTR~\citep{gao2023memotr} and SambaMOTR).
SambaMOTR sets a new state of the art, with $+3.8$ HOTA and $+5.2$ AssA on the strongest competitor MeMOTR.
Our method owes this performance improvement to its better modeling of the historical information, our effective strategy to learn accurate sequence models through occlusions, and our modeling of tracklets interaction (group dancers move synchronously).

\paragraph{BFT.} Bird flocks present similar appearance and non-linear motion. For this reason, OC-SORT works best among tracking-by-detection methods. Nevertheless, bird flocks move synchronously, and interaction among tracklets is an essential cue for modeling joint object motion. Thanks to our proposed sequence models synchronization, SambaMOTR achieves $+2.8$ HOTA and $+4.9$ AssA over the best competitor overall (OC-SORT), and an impressive $+6.3$ HOTA and $+12.5$ improvement over the previous best tracking-by-propagation method TrackFormer~\citep{meinhardt2022trackformer}.

\paragraph{SportsMOT.} Sports scenes typically present non-linear motion patterns that the Kalman filter struggles to model, hence the underwhelming performance of ByteTrack~\citep{zhang2022bytetrack}.
For this reason, trackers that model non-linear motion either explicitly (OC-SORT~\citep{cao2023observation}) or implicitly (TransTrack~\citep{sun2020transtrack}) perform well.
Notably, our tracking-by-propagation SambaMOTR enables implicit joint modeling of motion, appearance, and tracklet interaction, obtaining the best HOTA overall ($69.8$) despite the lower DetA of our Deformable-DETR detector compared to OC-SORT's YOLOX-X.
Moreover, SambaMOTR exhibits a significant $+1.6$ AssA over the best tracking-by-propagation method and an impressive $+4.6$ AssA over OC-SORT.

\begin{table}[t]
  \caption{\textbf{State-of-the-art comparison on SportsMOT}~\citep{cui2023sportsmot} without additional training data. Best tracking-by-propagation method in \textbf{bold}; best overall \underline{underlined}.}
  \label{tab:sportsmot}
  \centering
\tablesize
  \setlength{\tabcolsep}{5pt}{
    \begin{tabular}{llccccc}
      \toprule
      Methods & Detector  & HOTA & AssA & DetA & IDF1 & MOTA \\
      \midrule
      \bf \textit{Tracking-by-detection:} \\
      FairMOT~\citep{zhang2021fairmot}& CenterNet~\citep{duan2019centernet} & 49.3 & 34.7 & 70.2 & 53.5 & 86.4 \\
      \greyrule
      QDTrack~\citep{pang2021quasi} & \multirow{3}{*}{YOLOX-X~\citep{ge2021yolox}} & 60.4 & 47.2 & 77.5 & 62.3 & 90.1 \\
      ByteTrack~\citep{zhang2022bytetrack} & & 62.1 & 50.5 & 76.5 & 69.1 & 93.4 \\
      OC-SORT~\citep{cao2023observation} &  & 68.1 & 54.8 & \underline{84.8} & 68.0 & \underline{93.4} \\
      \greyrule
      TransTrack~\citep{sun2020transtrack} & Deformable DETR~\citep{zhu2020deformable} & 68.9 & 57.5 & 82.7 & 71.5 & 92.6 \\
      \midrule
      \bf \textit{Tracking-by-propagation:} \\
      MeMOTR~\citep{gao2023memotr} & \multirow{2}{*}{Deformable DETR~\citep{zhu2020deformable}} & 68.8 & 57.8 & 82.0 & 69.9 & 90.2 \\
      SambaMOTR (ours) &  & \bf \underline{69.8} & \bf \underline{59.4} & \bf 82.2 & \bf \underline{71.9} & \bf 90.3 \\
      \bottomrule
    \end{tabular}
  }
\end{table}

\begin{table}[htb]
    \small
    \setlength{\tabcolsep}{5pt}
    \caption{
        \textbf{Ablation on method components} on the DanceTrack test set. Compared to prior work (in {\grtext gray}), we introduce a long-range query propagation module based on state-space models (SSM), we mask uncertain queries during the state update (MaskObs), we synchronize memory representations across tracklets (Sync), and we learn from longer sequences (Longer).
    }
    \tablesize
    \label{tab:method_components}
    \begin{tabular}{lc|cccc|ccccc}
        \toprule
         Method & & SSM & MaskObs & Sync & Longer & HOTA & AssA & DetA & IDF1 & MOTA \\
        \midrule
        \multirow{4}{*}{SambaMOTR (\textbf{Ours})} & (a) & \checkmark & - & - & - & 63.5 & 53.8 & 75.1 & 67.0 & 81.7 \\
        & (b) & \checkmark & \checkmark & - & - & 64.8 & 54.3 & 77.7 & 68.1 & 85.7 \\
        & (c) & \checkmark & \checkmark & \checkmark & - & 65.9 & 55.6 & 78.4 & 68.7 & 87.4 \\
        & (d) & \checkmark & \checkmark & \checkmark & \checkmark & \bf 67.2 & \bf 57.5 & \bf 78.8 & \bf 70.5 & \bf 88.1 \\
        \midrule
        \grtext MOTR~\citep{zeng2022motr} & \grtext (e) & \grtext - & \grtext - & \grtext - & \grtext - & \grtext 54.2 & \grtext 40.2 & \grtext 73.5 & \grtext 51.5 & \grtext 79.7 \\ 
        \grtext MeMOTR~\citep{gao2023memotr} & \grtext (f) & \grtext - & \grtext - & \grtext - & \grtext - & \grtext 63.4 & \grtext 52.3 & \grtext 77.0 & \grtext 65.5 & \grtext 85.4 \\ 
        \bottomrule
    \end{tabular}
\end{table}

\subsection{Ablation Studies} \label{ssec:exp_ablation}
We ablate the effect of each component of our method in \cref{tab:method_components}, as detailed in \cref{sec:method} and illustrated in \cref{fig:method_components}. Additional ablation studies are presented in \cref{sec:app_results_ablations}.

\myparagraph{SSM.} Line (a) shows the benefits of a sequential representation for tracking. We use a vanilla sequence model, such as Mamba, as the baseline for query propagation, establishing a robust foundation that outperforms MeMOTR’s EMA-based history and temporal attention module.

\myparagraph{MaskObs.} Handling track queries during occlusions (line b) with MaskObs - which masks uncertain observations from the state update and relies on long-term memory and interactions with visible tracklets - leads to significant overall improvements ($+1.3$ HOTA), highlighting the effectiveness of managing occluded objects.

\myparagraph{Sync.} Making tracklets aware of each other through our synchronization mechanism (line c) results in over 1\% improvement across all metrics, demonstrating how modeling interactions between tracklets enhances tracking accuracy by capturing joint dynamics and coordinated movements.

\myparagraph{Long-sequence training.} Efficiently incorporating longer sequences during training (line d) helps the model to properly utilize its long-term memory, enabling generalization to indefinitely long sequences and leading to a notable $+1.9$ improvement in AssA.

Our final query propagation method (line d) improves MeMOTR's association accuracy by $+5.2$ (line f), and MOTR's by an impressive $+17.3$ (line e).

\section{Limitations} \label{sec:limitations}
Following the tracking-by-propagation paradigm, our model drops tracklets that are inactive for more than $N_{miss}$ frames to decrease the risk of ID switches. However, in some datasets like SportsMOT~\citep{cui2023sportsmot} football players may disappear from the camera view for multiple seconds, outliving the $N_{miss}$ threshold.
We argue that future work should complement tracking-by-propagation with long-term re-identification to tackle this issue.
Furthermore, in this paper, we introduced Samba, a set-of-sequences model. Our ablation study (\cref{tab:method_components}) shows that Samba significantly outperforms the already strong \ac{ssm} baseline. However, this comes with the trade-off of increased computational complexity. In particular, \acp{ssm} have linear complexity in time and linear complexity in the number of sequences (tracklets) independently modeled. Samba retains linear-time complexity, which enables it to track for indefinitely long-time horizons, but quadratic complexity in the number of sequences due to the use of self-attention in memory synchronization. Our ablations show that this trade-off is worth the performance improvement.

\section{Conclusion} \label{sec:conclusions}
The proposed SambaMOTR fully leverages the sequential nature of the tracking task by using our set-of-sequences model, Samba, as a query propagation module to jointly model the temporal history of each tracklet and their interactions.
The resulting tracker runs with linear-time complexity and can track objects across indefinitely long sequences.
SambaMOTR surpasses the state-of-the-art on all benchmarks, reporting significant improvements in association accuracy compared to prior work.

\clearpage

\subsubsection*{Acknowledgments}
This work was supported in part by the Max Plank ETH Center for Learning Systems.
\bibliography{iclr2025_conference}
\bibliographystyle{iclr2025_conference}
\clearpage


\appendix

\section*{Appendix}
\def\thesection{\Alph{section}}
\setcounter{section}{0}
\setcounter{table}{0}
\renewcommand{\thetable}{\Alph{table}}
\setcounter{figure}{0}
\renewcommand{\thefigure}{\Alph{figure}}


In this appendix, we report additional discussions and experiments.  
First, we provide background on sequence models in \cref{app:sequence_models}.
Then, we report additional implementation details for SambaMOTR in \cref{sec:app_sambamotr}.
We show a schematic illustration of the Samba block in \cref{fig:samba} and our method components in \cref{fig:method_components}.
Finally, we provide additional results \cref{sec:app_results}, 
conducting several ablation studies on specific design choices that contributed to SambaMOTR's performance.

\section{Background on Sequence Models.} \label{app:sequence_models}
\paragraph{Sequence Models.}
Sequence models are a class of machine learning models dealing with sequential data, \ie where the order of elements is important.
Applications of sequence models are widespread across different fields, such as natural language processing~\citep{vaswani2017attention,gu2023mamba}, time series forecasting~\citep{wen2022transformers} and video analysis~\citep{venugopalan2015sequence}.
Several architectures have been proposed to process sequences, each with its own strengths and limitations.
\textit{\Acp{rnn}} handles sequential data by maintaining a hidden state that updates as the network processes each element in a sequence. However, RNNs often struggle with long sequences due to issues like vanishing or exploding gradients~\citep{pascanu2013difficulty}.
\textit{\Ac{lstm}} networks~\citep{hochreiter1997long} introduce gating units to mitigate \ac{rnn}'s vanishing gradient problem.
\textit{Transformers}~\citep{vaswani2017attention} rely on self-attention mechanisms to weigh the importance of different parts of the input data. Unlike \acp{rnn} and \acp{lstm}, transformers process entire sequences simultaneously, making them efficient at modeling long-range dependencies at the cost of quadratic computational complexity wrt. sequence length.
Building on the idea of modeling temporal dynamics like \acp{rnn} and \acp{lstm}, structured state-space models~\citep{gu2021efficiently} introduce a principled approach to state management inspired by classical \acp{ssm}~\citep{kalman1960new}. Despite excelling at modeling long-range dependencies in continuous signals, structured \acp{ssm} lag behind transformers on discrete modalities such as text. Recently, selective state-space models (Mamba)~\citep{gu2023mamba} improved over prior work by making the \ac{ssm} parameters input-dependent, achieving the modeling power of Transformers while scaling linearly with sequence length.

\paragraph{Set-of-sequences Models.}
Only few approaches~\citep{yang2017deep,amiridi2022latent,wu2024feasibility} explore the task of set-of-sequences modeling, which we define as the task of simultaneously modeling multiple temporal sequences and their interdependencies to capture complex relationships and interactions across different data streams. 
Set-of-sequences modeling has applications in multivariate time series analysis~\citep{amiridi2022latent}, dynamic graph modeling~\citep{wu2024feasibility}, and sensor data fusion~\citep{yang2017deep}.
However, existing techniques involve complex and expensive designs. 
We here introduce Samba, a linear-time set-of-sequences model based on the synchronization of multiple selective state-space models to account for the interaction across sequences.
%

%
\section{SambaMOTR - Additional Details} \label{sec:app_sambamotr}

\begin{figure}
  \centering
  \includegraphics[width=1.0\linewidth,trim={0.0cm 0.0cm 0.0cm 0.0cm},clip]{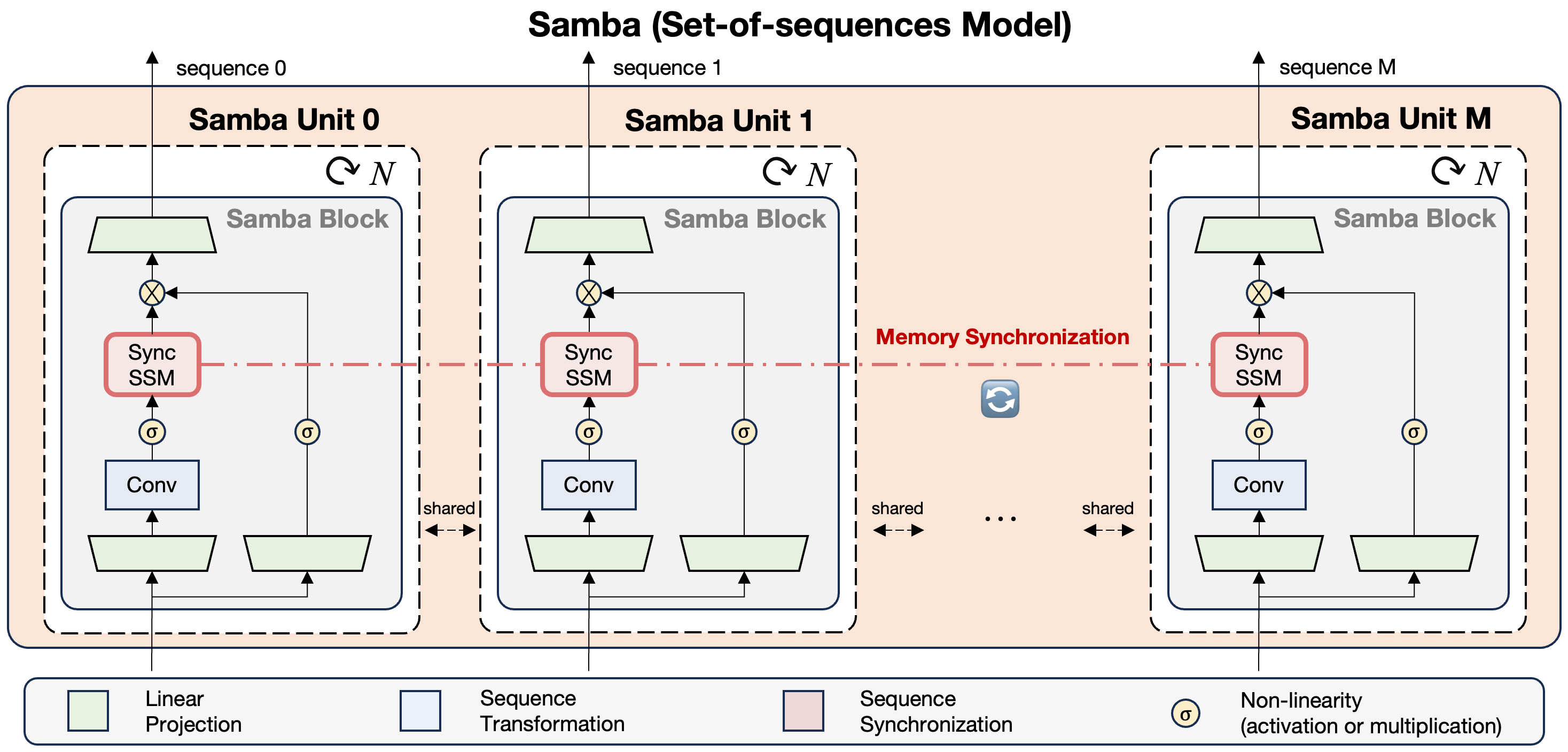}
  \caption{\textbf{Illustration of our Set-of-sequences Model block.} Our set-of-sequences model Samba simultaneously processes an arbitrary number $M$ of input sequences. Each sequence is processed by a Samba unit, synchronized with the others thanks to our synchronized state-space model. All Samba units share weights and are composed of a stack of $N$  Samba blocks. A Samba block has the same architecture as a Mamba block, but it adopts our synchronized SSM to synchronize long-term memory representations across the individual state-space models.} \label{fig:samba}
\end{figure}

\begin{figure}[ht]
  \centering
  \small
  \begin{tabular}{cc}
    \toprule
     & Schematic Illustration of our Contributions \\
    \midrule
    \raisebox{+1.3\normalbaselineskip}[0pt][0pt]{\rotatebox[origin=c]{90}{\makecell{SSM}}} & \includegraphics[width=0.92\linewidth,trim={2.25cm 16.1cm 1.0cm 0.0cm},clip]{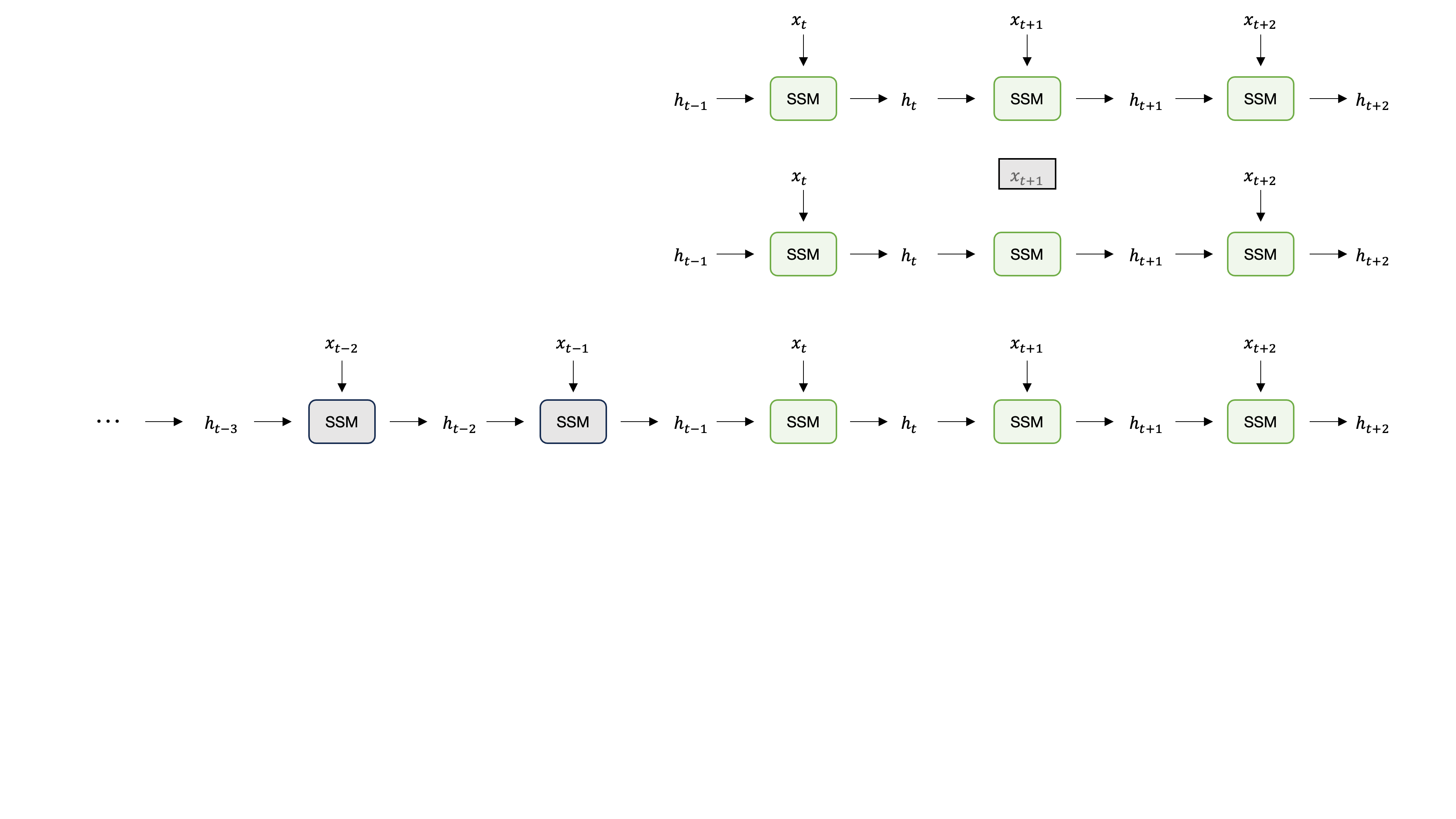} \\
    \midrule
    \raisebox{+1.4\normalbaselineskip}[0pt][0pt]{\rotatebox[origin=c]{90}{MaskObs}}  & \includegraphics[width=0.92\linewidth,trim={2.25cm 12.5cm 1.0cm 3.5cm},clip]{figures/images/components.png} \\
    \midrule
    \raisebox{+6.8\normalbaselineskip}[0pt][0pt]{\rotatebox[origin=c]{90}{\makecell{Sync}}} & \includegraphics[width=0.92\linewidth,trim={2.25cm 4.8cm 1.0cm 1.3cm},clip]{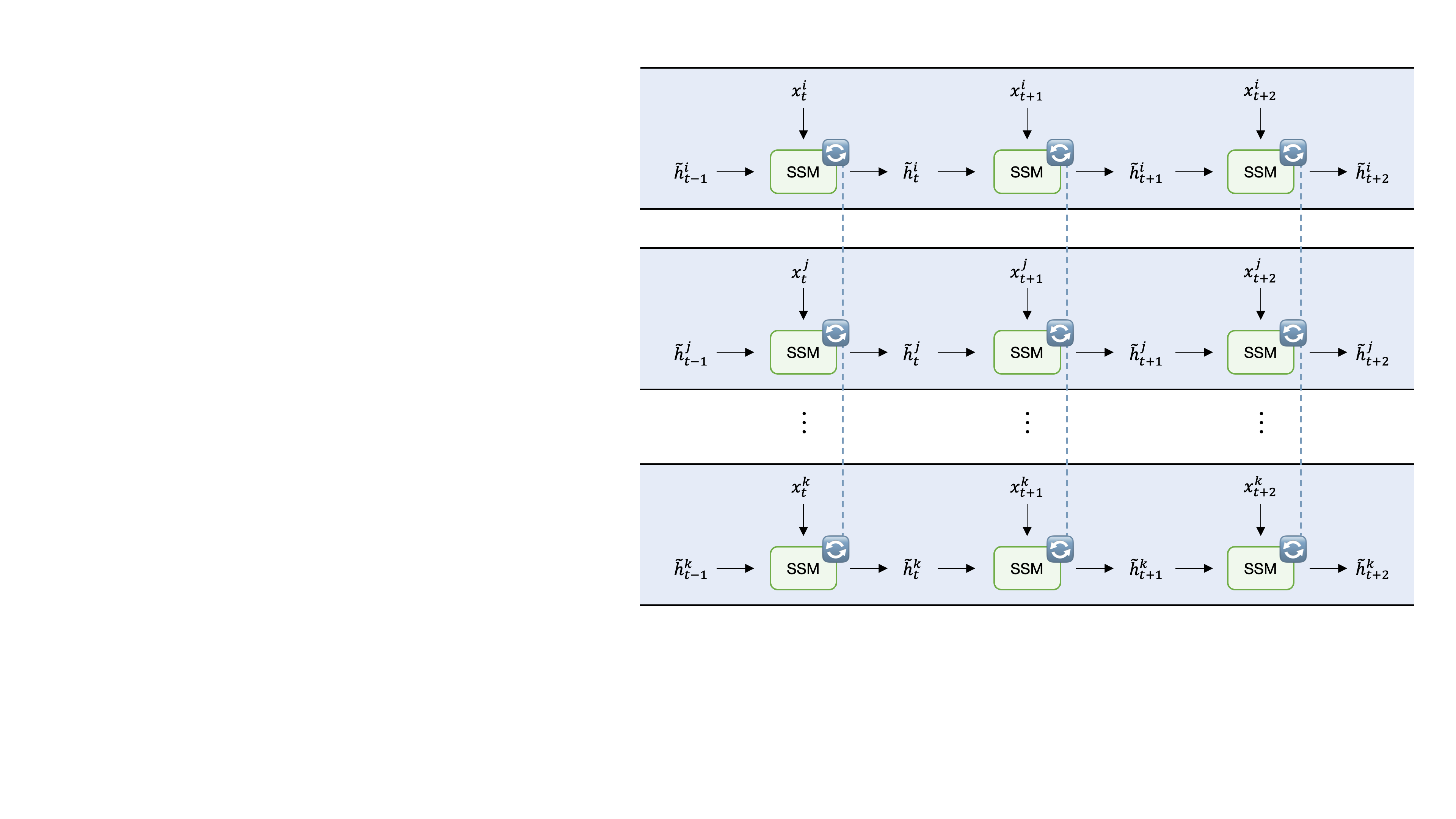} \\
    \midrule
    \raisebox{+1.3\normalbaselineskip}[0pt][0pt]{\rotatebox[origin=c]{90}{\makecell{Longer}}} & \includegraphics[width=0.92\linewidth,trim={2.25cm 8.6cm 1.0cm 7.5cm},clip]{figures/images/components.png} \\
    \bottomrule
  \end{tabular}
  \caption{Schematic illustration of our contributions (as ablated in \cref{tab:method_components}). State-space model (SSM) blocks at timesteps with gradient applied are in green, and blocks without gradient are in grey.}
  \label{fig:method_components}
\end{figure}

SambaMOTR builds on Samba to introduce linear-time sequence modeling in tracking-by-propagation, treating each tracklet as a sequence of queries and autoregressively predicting the future track query.
By inducing synchronization on the SSMs' memories across an arbitrary number of sequences, Samba elegantly models tracklet interaction and query propagation under occlusions.

\subsection{Samba} \label{sec:app_samba}
We illustrate a Samba set-of-sequences model in \cref{fig:samba}. 
A Samba model (\cref{fig:samba}) is composed of a set of siamese Samba units (one for each sequence being modeled) with shared weights. 
Each Samba unit is synchronized with others through our synchronized \ac{ssm} layer. 
In particular, a Samba unit is composed of $N$ non-linear Samba blocks.
To obtain a non-linear Samba unitblock that can be embedded into a neural network, we wrap the synchronized \ac{ssm} layer following the Mamba~\citep{gu2023mamba} architecture.
A linear projection expands the input dimension $D$ by an expansion factor $E$, followed by a causal convolution and a SiLU~\citep{hendrycks2016gaussian} activation before being fed to the sync \ac{ssm} layer. The output of a residual connection is passed to a SiLU before being multiplied by the output of the synchronized SSM and passed to an output linear projection.
Moreover, we replace Mamba's RMSNorm~\citep{zhang2019root} with LayerNorm~\citep{ba2016layer} for consistency with the detector.
Finally, we repeat $N$ such Samba blocks, interleaved with standard normalization and residual connections, to form a Samba unit. 
The resulting set-of-sequences model is linear-time, supports a variable number of sequences that start and end at different time steps, and models long-range relationships and interdependencies across multiple sequences.
\subsection{Schematic Illustration of Our Contributions}
We provide a schematic illustration of our contributions towards building Samba in \cref{fig:method_components}, disentangling them from one another to make the functioning of each component clear.

Mamba is the underlying sequence model, shown in the first row (Mamba).
The second row depicts our strategy to deal with uncertain observations by ignoring them in the state update (Occlusion Masking).
Synchronization across multiple sequence models using our synchronization module to let their hidden states communicate to model sequence interaction is shown in the third row (Sync).
The last row illustrates our efficient training strategy to learn long-range dynamics from longer sequences at a comparable computational expense for backpropagation (Longer).

Each of these components is ablated in \cref{tab:method_components} by incrementally adding them within our framework, showing the effectiveness of each towards the impressive final performance of SambaMOTR.

\section{Additional Results} \label{sec:app_results}

\begin{table}[t]
  \caption{
  State-of-the-art comparison on MOT17~\citep{milan2016mot16}. Best tracking-by-propagation method in \textbf{bold}; best overall \underline{underlined}. For a fair comparison with the only MeMOTR's published result, we also adopt DAB-Deformable-DETR.
  }
  \label{tab:mot17}
  \centering
  \tablesize
    \setlength{\tabcolsep}{3pt}{\begin{tabular}{llccccc}
      \toprule
      Methods & Detector & HOTA & AssA & DetA & IDF1 & MOTA \\
      \midrule
      \textit{Tracking-by-detection:} \\
          CenterTrack~\citep{zhou2020tracking} & \multirow{2}{*}{CenterNet~\citep{duan2019centernet}} & 52.2 & 51.0 & 53.8 & 64.7 & 67.8 \\
          FairMOT~\citep{zhang2021fairmot} &  & 59.3 & 58.0 & 60.9 & 72.3 & 73.7 \\
          \greyrule
          TransTrack~\citep{sun2020transtrack} & \multirow{3}{*}{Deformable DETR~\citep{zhu2020deformable}} & 54.1 & 47.9 & 61.6 & 63.9 & 74.5 \\
          TransCenter~\citep{xu2022transcenter} &  & 54.5 & 49.7 & 60.1 & 62.2 & 73.2 \\
          MeMOT~\citep{cai2022memot} &  & 56.9 & 55.2 & - & 69.0 & 72.5 \\
          \greyrule
          GTR~\citep{zhou2022global} & CenterNet2~\citep{zhou2021probabilistic} & 59.1 & 57.0 & 61.6 & 71.5 & 75.3 \\
          \greyrule
          DeepSORT~\citep{wojke2017simple} & \multirow{7}{*}{YOLOX-X~\citep{ge2021yolox}}  & 61.2 & 59.7 & 63.1 & 74.5 & 78.0 \\
          SORT~\citep{bewley2016simple} &  & 63.0 & 62.2 & 64.2 & 78.2 & 80.1 \\
          ByteTrack~\citep{zhang2022bytetrack} &  & 63.1 & 62.0 & 64.5 & 77.3 & 80.3 \\
          OC-SORT~\citep{cao2023observation} &  & 63.2 & 63.4 & 63.2 & 77.5 & 78.0 \\
          QDTrack~\citep{pang2021quasi} &  & 63.5 & 62.6 & 64.5 & 77.5 & 78.7 \\
          C-BIoU~\citep{yang2023hard} &  & 64.1 & 63.7 & 64.8 & 79.7 & 81.1 \\
          MotionTrack~\citep{qin2023motiontrack} &  & 65.1 & 65.1 & 65.4 & 80.1 & 81.1 \\
      \midrule
      \textit{Tracking-by-propagation:} \\
          TrackFormer~\citep{meinhardt2022trackformer} & \multirow{2}{*}{Deformable DETR~\citep{zhu2020deformable}}  & - & - & - & 68.0 & 74.1 \\
          MOTR~\citep{zeng2022motr} &  & 57.2 & 55.8 & 58.9 & 68.4 & 71.9 \\
          \greyrule
          MeMOTR~\citep{gao2023memotr} & \multirow{2}{*}{DAB-Deformable DETR~\citep{liu2022dab}} & 58.8 & 58.4 & 59.6 & 71.5 & 72.8 \\
          SambaMOTR (ours) & & 58.8 & 58.2 & 59.7 & 71.0 & 72.9 \\
      \bottomrule
    \end{tabular}}
\end{table}

We report additional results on the popular MOT17 pedestrian tracking benchmark in \cref{sec:app_results_mot17}.
We extend our ablation study in \cref{sec:app_results_ablations}, investigating the effectiveness of synchronization, the use of positional embeddings and the effectiveness of residual prediction.

\subsection{MOT17}  \label{sec:app_results_mot17}
While MOT17 served as a benchmark of paramount importance to advance the state of current multiple object tracking algorithms, its very small size is reducing its significance as a training dataset. Since MOT17 only counts 7 training videos, modern tracking solutions complement its training with additional detection datasets and increasingly stronger detectors to improve the overall tracking performance and top the leaderboard. However, such expedients are deviating from the study of fundamental tracking solutions and focusing more on engineering tricks. Moreover, due to its highly linear motion, its small size (only 7 videos), and the subsequent need for training on additional detection datasets, end-to-end tracking methods do not provide additional advantages over more naive Kalman-filter-based methods. For this reason, we preferred to it other more modern and meaningful datasets in the main paper, \ie DanceTrack~\citep{sun2022dancetrack}, SportsMOT~\citep{cui2023sportsmot} and BFT~\citep{zheng2024nettrack}, which allows us to study the importance of modeling tracklets interaction and of implicitly learning motion and appearance models to cope with the underlying non-linear motion, appearance and pose changes of the objects.
Nevertheless, we here compare with the state-of-the-art for completeness and show comparable performance to previous tracking-by-propagation methods.

\subsection{Ablation Studies} \label{sec:app_results_ablations}
We here complement the ablation study in \cref{ssec:exp_ablation} with additional experiments on specific SambaMOTR's design choices. All ablations are based on the final version of our method, including all contributions as in \cref{tab:method_components} line d.

\paragraph{Ablation on different formulations of synchronization.} In \cref{tab:state_synchronization}, we ablate on different formulations of state synchronization and report the corresponding state update equation for each option. In particular, the first row (Sync: -) does not apply state synchronization and is equivalent to using Mamba as a query propagation module together with our occlusion masking and efficient longer training strategy as explained in \cref{ssec:sambamotr_propagation}. Since this option does not model tracklet interaction, it reports the lowest performance.
We then compare synchronizing the hidden state before (prior) or after (posterior) and find that synchronizing the posterior is more effective. We attribute this to the opportunity to compensate for occluded observations in the current frame with the dynamics from other unoccluded tracklets to better model track query propagation through occlusions.

\paragraph{Ablation on the effect of synchronization on hard DanceTrack sequences.} 
In \cref{tab:hard}, we report the performance on hard sequences of the DanceTrack test set for two SambaMOTR with and without synchronization. We select the top-6 hardest sequences for the version without synchronization and show that utilizing synchronization greatly improves the overall metrics.

\paragraph{Ablation on the use of query positional embeddings in Samba.} 
In \cref{tab:query_pos}, we ablate on the addition of positional embeddings to the track embeddings before feeding them to Samba. We find that positional embeddings are very beneficial to Samba, arguably because they enable to implicitly learn non-linear motion models.

\paragraph{Ablation on the prediction of residual vs. full queries with Samba.} 
In \cref{tab:residual}, we ablate on the output format of our Samba-based query propagation module. We compare two versions: one that directly outputs the final track queries with Samba, and one that predicts a residual over the track queries used to detect in the current frame. We find that learning a residual is significantly more effective than directly predicting the final track query.

\paragraph{Ablation on the query propagation strategy through occlusions.} 
We compare two query propagation strategies for occluded track queries in \cref{tab:occlusions}. First, we evaluate our model with MeMOTR’s~\citep{gao2023memotr} query propagation strategy (Freeze), which freezes the last observed state - \ie the last track query that generated a confident detection - and memory until the tracklet is detected again in a new frame.
Next, we compare this with actively propagating occluded track queries and their memory through occlusions using our MaskObs strategy (\cref{ssec:sambamotr_propagation}).
We find that MaskObs outperforms Freeze: by inferring a tracklet’s future state during occlusions using only its past memory and interactions with other observed objects, it keeps tracklets alive longer.

\begin{table}[htb]
  \caption{\textbf{Ablation on the effect of synchronization} on difficult sequences on DanceTrack test.}
  \label{tab:hard}
  \tablesize
  \begin{center}
      \begin{tabular}{c|ccc|ccc}
      \toprule
      \multirow{3}{*}{Sequence} & \multicolumn{6}{c}{Synchronization} \\
      \cmidrule{2-7}
       & \multicolumn{3}{c|}{\xmark} & \multicolumn{3}{c}{\checkmark} \\
      \cmidrule{2-7}
       & HOTA & DetA & AssA & HOTA & DetA & AssA \\
      \midrule
      dancetrack0046 &	34.5	&54.0	&22.1 & 39.8	&60.7&	26.2  \\
      dancetrack0085&	39.6	&60.9	&25.9 & 40.3	&63.9	&25.5  \\
      dancetrack0050	&41.9	&66.4&	26.5 & 42.4&	69.8&	25.8  \\
      dancetrack0036&	42.9	&74.7	&24.7 & 48.4	&77.5&	30.3  \\
      dancetrack0028	&43.1	&71.9&	25.8 & 47.8	&73.6	&31.1  \\
      dancetrack0009	&43.4&	73.6	&25.7  &48.0	&75.2	& 30.7  \\
      \midrule
      \textit{average} & 40.9  & 66.9  & 25.1  & \bf 44.5 & \bf 70.1 & \bf 28.3  \\
      \bottomrule  
      \end{tabular}
  \end{center}
\end{table}

\begin{table}[htb]
\tablesize
\caption{\textbf{Ablation on memory synchronization positioning.} We report the state equation and performance on the DanceTrack test set for: (i) the baseline without synchronization (-); (ii) synchronization on the updated state prior to input contribution (Prior); (iii) synchronization on the fully-updated state (Posterior).}
\label{tab:state_synchronization}
\centering 
\begin{tabular}{@{}llccccc@{}}
\toprule
Sync & State Equation                                                                                                               & HOTA          & AssA & DetA          & IDF1          &     MOTA      \\ \midrule
 - & $h_t^i = \mathbf{\bar{A}^i}(t)h^i_{t-1} + \mathbf{\bar{B}^i}(t)x^i_t$                                          & 66.0              & 56.4     & 77.5              &       69.5        &  86.7             \\
 Prior & $h_t^i = \Gamma_{i \in \mathcal{T}}\left(\mathbf{\bar{A}^i}(t)h^i_{t-1}\right) + \mathbf{\bar{B}^i}(t)x^i_t$  &  66.1             & 56.7     & 77.3               & 70.0               & 86.4               \\
 Posterior & $h_t^i = \Gamma_{i \in \mathcal{T}}\left(\mathbf{\bar{A}^i}(t)h^i_{t-1} + \mathbf{\bar{B}^i}(t)x^i_t\right)$ & \bf   67.2           & \bf 57.5      & \bf  78.8               &  \bf     70.5          &  \bf  88.1     \\ \bottomrule
\end{tabular}
\end{table}
\begin{table}[htb]
\tablesize
\caption{\textbf{Ablation on the use of positional embeddings.} We ablate on the addition of positional embeddings to the observed queries fed as input to the Samba module.}
\label{tab:query_pos}
\centering 
\begin{tabular}{@{}cccccc@{}}
\toprule
Query Position   &           HOTA          & AssA & DetA          & IDF1          &     MOTA      \\ \midrule
 - &     65.6 & 56.2 & 76.7 & 69.3 & 85.4 \\           
 \checkmark &  \bf   67.2           & \bf 57.5      & \bf  78.8               &  \bf     70.5          &  \bf  88.1     \\ \bottomrule
\end{tabular}
\end{table} 
\begin{table}[htb]
\tablesize
\caption{\textbf{Ablation on the use of residual prediction.} We evaluate two formats for the output of SambaMOTR's Samba module, \ie direct query prediction (-) and prediction of a residual wrt. the track query from the previous frame (\checkmark).}
\label{tab:residual}
\centering 
\begin{tabular}{@{}cccccc@{}}
\toprule
Residual   &           HOTA          & AssA & DetA          & IDF1          &     MOTA      \\ \midrule
 -            &  64.2    & 54.0              &  76.7      & 67.0       &    84.5           \\
 \checkmark &  \bf   67.2           & \bf 57.5      & \bf  78.8               &  \bf     70.5          &  \bf  88.1     \\ \bottomrule
\end{tabular}
\end{table}  
\begin{table}[htb]
\tablesize
\caption{\textbf{Ablation on strategies for tracking through occlusions.} We evaluate two strategies for tracking objects through occlusions: freezing the last observed track state (Freeze) as in MeMOTR, and propagating queries through occlusions using our MaskObs strategy.}
\label{tab:occlusions}
\centering 
\begin{tabular}{@{}lccccc@{}}
\toprule
Strategy   &           HOTA          & AssA & DetA          & IDF1          &     MOTA      \\ \midrule
 Freeze            &  65.9    & 56.6              &  76.8      & 69.7       &    86.3           \\
 MaskObs &  \bf   67.2           & \bf 57.5      & \bf  78.8               &  \bf     70.5          &  \bf  88.1     \\ \bottomrule
\end{tabular}
\end{table}  

\end{document}